\documentclass[letterpaper]{article} % DO NOT CHANGE THIS
\usepackage{aaai25}  % DO NOT CHANGE THIS
\usepackage{times}  % DO NOT CHANGE THIS
\usepackage{helvet}  % DO NOT CHANGE THIS
\usepackage{courier}  % DO NOT CHANGE THIS
\usepackage[hyphens]{url}  % DO NOT CHANGE THIS
\usepackage{graphicx} % DO NOT CHANGE THIS
\usepackage{natbib}  % DO NOT CHANGE THIS AND DO NOT ADD ANY OPTIONS TO IT
\usepackage{caption} % DO NOT CHANGE THIS AND DO NOT ADD ANY OPTIONS TO IT
\usepackage{boldline,multirow,booktabs, subfig}
\newcommand{\shortname}{LAMA-UT}
\usepackage{algorithm}
\usepackage{algorithmic}
\usepackage{pgfplots}
\pgfplotsset{compat=1.18}

\usepackage{newfloat}
\usepackage{listings}
\DeclareCaptionStyle{ruled}{labelfont=normalfont,labelsep=colon,strut=off} % DO NOT CHANGE THIS
\floatstyle{ruled}
\newfloat{listing}{tb}{lst}{}
\floatname{listing}{Listing}
\title{LAMA-UT: Language Agnostic Multilingual ASR through Orthography Unification and Language-Specific Transliteration}
\author{
    Sangmin Lee\textsuperscript{\rm 1},
    Woojin Chung\textsuperscript{\rm 1},
    Hong-Goo Kang\textsuperscript{\rm 1}\\
}
\affiliations{
    \textsuperscript{\rm 1}Dept. of Electrical \& Electronic Engineering, Yonsei University, South Korea\\
    \{sangmin\_lee, woojinchung\}@dsp.yonsei.ac.kr, hgkang@yonsei.ac.kr
}

\begin{document}

\maketitle

\begin{abstract}

Building a universal multilingual automatic speech recognition (ASR) model that performs equitably across languages has long been a challenge due to its inherent difficulties.
To address this task we introduce a \textbf{L}anguage-\textbf{A}gnostic \textbf{M}ultilingual \textbf{A}SR pipeline through orthography \textbf{U}nification and language-specific \textbf{T}ransliteration (\shortname\footnote{Codes: https://github.com/sanghyang00/lama-ut}).
\shortname~operates without any language-specific modules while matching the performance of state-of-the-art models trained on a minimal amount of data.
Our pipeline consists of two key steps.
First, we utilize a universal transcription generator to unify orthographic features into Romanized form and capture common phonetic characteristics across diverse languages. 
Second, we utilize a universal converter to transform these universal transcriptions into language-specific ones.
In experiments, we demonstrate the effectiveness of our proposed method leveraging universal transcriptions for massively multilingual ASR.
Our pipeline achieves a relative error reduction rate of 45\% when compared to Whisper and performs comparably to MMS, despite being trained on only 0.1\% of Whisper's training data.
Furthermore, our pipeline does not rely on any language-specific modules.
However, it performs on par with zero-shot ASR approaches which utilize additional language-specific lexicons and language models.
We expect this framework to serve as a cornerstone for flexible multilingual ASR systems that are generalizable even to unseen languages.

\end{abstract}
\section{Introduction}

Developing a model for multilingual automatic speech recognition (ASR) is appealing due to its applicability to universal languages, including low-resource or unseen languages.
However, this task presents significant challenges as it requires extensive datasets and involves the complexity of capturing shared characteristics across diverse languages in both phonetic and orthographic domains.

Since the revolution in ASR technologies driven by self-supervised learning (SSL) models~\cite{NEURIPS2020_92d1e1eb, conneau2020unsupervisedcrosslingualrepresentationlearning, hsu2021hubert, chen2022wavlm}, monolingual ASR has achieved superhuman transcription performance, shifting the main focus of recent research towards developing a universal model that spans multiple languages.

There are two primary methods for building a multilingual ASR model.
The first approach involves scaling the size of both the labeled dataset and the model itself, using a single universal model to enhance its capacity and cover a vast number (100+) of languages, thereby achieving multilingual ASR~\cite{radford2023robust}.
Another approach involves incorporating language-specific modules into the universal model to address the performance inconsistencies of previous methods. 
For example, MMS~\cite {JMLR:v25:23-1318} demonstrated the feasibility of scaling multilingual technology to over 1,000 languages by leveraging common features across languages and adding language-specific modules to improve the performance of each language.
Indeed, there have been efforts to integrate both methods~\cite{zhang2023googleusmscalingautomatic}, combining their strengths to build a more robust and versatile model.

Although these works have demonstrated strong performance across various languages, the trade-off between performance and complexity remains a substantial challenge.
The first method, using a single universal pipeline, struggles to achieve consistent performance across languages, and its effectiveness in low-resource languages remains uncertain.
On the other hand, despite achieving state-of-the-art performance and parameter efficiency, the second method cannot be considered a single universal model due to the inclusion of language-specific modules. Moreover, the use of language-specific modules like adapters, heads, and language models (LMs) sometimes complicates the training and inference pipeline, suggesting potential areas for future improvement.
\begin{figure*}[]
    \centering
    \includegraphics[width=\textwidth]{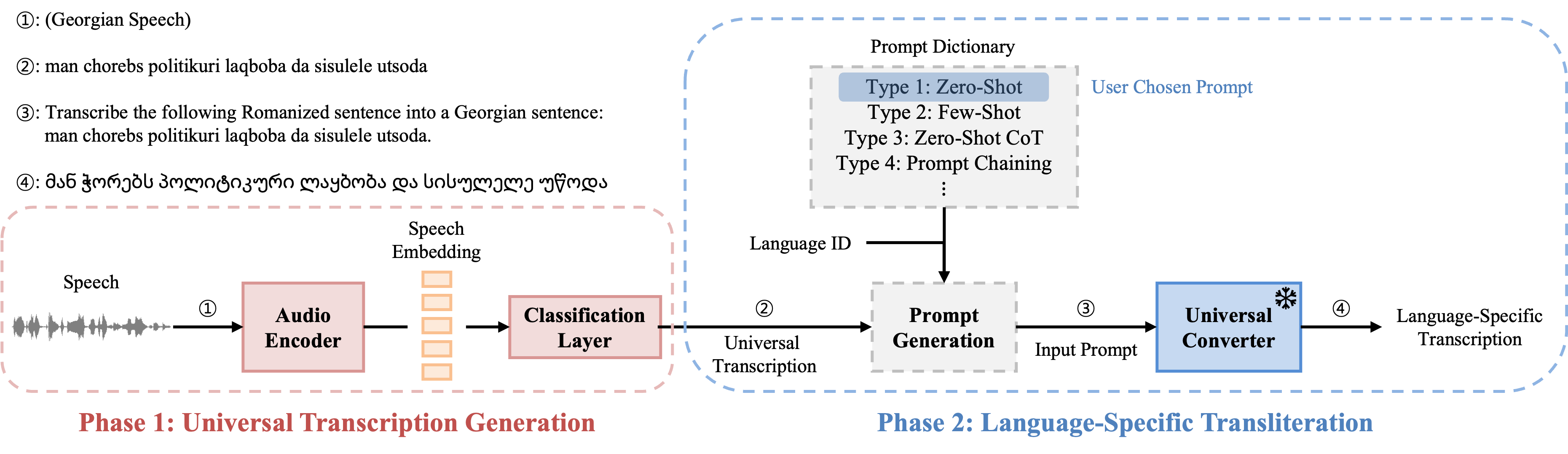}
    \caption{Illustration of our universal ASR pipeline.}
    \label{fig:pipeline}
\end{figure*}

Simultaneously, large language models (LLMs) have garnered considerable attention for their remarkable capabilities in the natural language processing (NLP) domain.
Following this trend, ASR pipelines have integrated audio SSL models as encoders and LLMs as decoders to enhance transcription quality~\cite{li2023prompting, fathullah2024prompting}.
These approaches involve using projectors~\cite{yu2024connecting} or fine-tuning strategies~\cite{tang2024salmonn, du2024lauragptlistenattendunderstand} to align modalities and improve transcription capabilities across multilingual datasets.
Subsequently, shallow fusion~\cite{chorowski2016betterdecodinglanguagemodel, kannan2018analysis} based scoring methods~\cite{hu2023massively, huang2024multilingual} were attempted to replace conventional LMs with LLMs during the decoding stage.
Despite these efforts resulting in performance gain across various languages, a comprehensive method to fully leverage the diverse emergent abilities of LLMs remains to be developed.

In this paper, we introduce a novel language-agnostic multilingual ASR pipeline that spans over 100 languages, including completely unseen languages. 
As in Fig~\ref{fig:pipeline}, the proposed pipeline consists of two phases: universal transcription generation and language-specific transliteration.
In the universal transcription generation phase, we focused on reducing orthographic complexity by unifying diverse orthographic systems into a consistent format, approximating phonetic features across multiple languages.
In the language-specific transliteration phase, we regard the transformation from universal transcription to language-specific transcription as a transliteration task by leveraging a universal converter.
Our experiments demonstrate notable transcription performance of~\shortname~across over 100 languages while using significantly smaller training data (only 680 hours) compared to other state-of-the-art multilingual ASR models.
Furthermore, our proposed pipeline outperforms previous methods, especially in low-resource languages, and demonstrates proficiency in completely unseen languages, achieving performance comparable to existing language-agnostic ASR methods without relying on any language-specific modules.
Our contributions are summarized as follows:

\begin{itemize}
    \item We propose a novel language-agnostic multilingual ASR pipeline consisting of two phases: universal transcription generation and language-specific transliteration.
    \item We enabled our proposed pipeline to perform multilingual ASR with minimal data by unifying diverse orthographic systems through Romanization.
    \item Our pipeline demonstrates consistent performance across over 100 seen languages and excels with completely unseen languages, all without relying on any language-specific modules or additional fine-tuning.
    
\end{itemize}
\section{Related Works}
\subsection{Multilingual ASR}
Initially, multilingual ASR models handled a limited number of languages~\cite{toshniwal2018multilingual, pratap2020massivelymultilingualasr50}, under 60. 
However, recent advancements have led to the development of models capable of managing a broader range of languages.
Whisper~\cite{radford2023robust} uses a sequence-to-sequence~\cite{NIPS2014_a14ac55a} approach with 680,000 hours of weakly supervised data, and its neural decoder serves as a LM, enhancing transcription performance.
With this method, Whisper attained impressive performance across most supported languages.

Google USM~\cite{zhang2023googleusmscalingautomatic} employs a Conformer~\cite{gulati20_interspeech} encoder with various types of heads~\cite{graves2012sequencetransductionrecurrentneural, chan2016listen} and is trained on an extensive dataset. 
It also employs a three-stage training incorporating speech-only, speech-text paired, and text-only data. 
Furthermore, to enhance transcription performance for low-resource languages, USM integrates language-specific adapters and employs Noisy Student Training (NST) techniques~\cite{9156610, Park_2020}. 

MMS~\cite{JMLR:v25:23-1318}, a state-of-the-art multilingual ASR model, employed a Connectionist Temporal Classification (CTC) based approach~\cite{graves2006connectionist} on a dataset covering over 1,000 languages. 
It utilizes a two-stage fine-tuning pipeline.
The first stage involves Romanization-based fine-tuning to learn a global representation across diverse languages. 
In the second stage, language-specific adapters and heads are added to capture detailed features for each language and fine-tuned.

\subsection{Zero-Shot ASR}
ASR-2K~\cite{li22aa_interspeech} is a zero-shot ASR model which utilizes three universal models to cover a range of languages: an acoustic model~\cite{li2020universal}, a pronunciation model~\cite{li2022zero}, and a LM~\cite{scannell12007crubadan}.
This suggests the potential for a universal multilingual ASR model capable of functioning in a zero-shot environment without relying on any language-specific components.
Consequently, Zero-Shot MMS~\cite{zhao2024scalingsimpleapproachzeroshot} utilized language-specific lexicon and n-gram LMs in the decoding phase to enhance zero-shot transcription performance. 

\subsection{LLM-Supported Multilingual ASR}
\citealt{hu2023massively} trained a multilingual LLM covering 84 languages and employed a shallow fusion-based per-frame scoring to enhance transcription quality in multilingual ASR.
Subsequently,~\citealt{huang2024multilingual} introduced non-autoregressive per-segment scoring, which improves transcription performance and reduces the computational burden.
These methods primarily leveraged the strengths of a multilingual acoustic model (USM) and achieved further accuracy by incorporating LLMs into the decoding step.
\section{Proposed Method}
The overall structure of the proposed multilingual ASR pipeline,~\shortname, comprises a universal transcription generation phase and a language-specific transliteration phase, as shown in Fig~\ref{fig:pipeline}. 
We produce universal transcriptions by finetuning an audio encoder with an additional classification layer.
Consequently, we manually select a prompt type from a predefined dictionary and combine it with language information to generate the input prompt for the universal converter.
Finally, by feeding this input prompt into the universal converter, we translate the universal transcription into language-specific ones.

\subsection{Universal Transcription Generation}
Previous studies in linguistics~\cite{ladefoged1996sounds, clark2007introduction} have shown that the phonological characteristics of human speech are constrained by a limited range of sounds due to the anatomical structure of the vocal tract. 
Similarly, in the ASR domain, prior research~\cite{taguchi-chiang-2024-language} has empirically demonstrated that the primary obstacle in multilingual ASR is the orthographic complexity across languages.
Through the integration of these two insights, we aim to unify orthographic systems across diverse languages by standardization of notations into a Latin character system.
This approach establishes alignment between phonetic and orthographic features through a unified transcription system.
As a result, we develop a universal transcription generator capable of producing consistent transcriptions across multilingual speech corpora, including unseen languages. 

\subsubsection{International Phonetic Alphabet.} 
\begin{figure}[]
    \centering
    \includegraphics[width=\columnwidth]{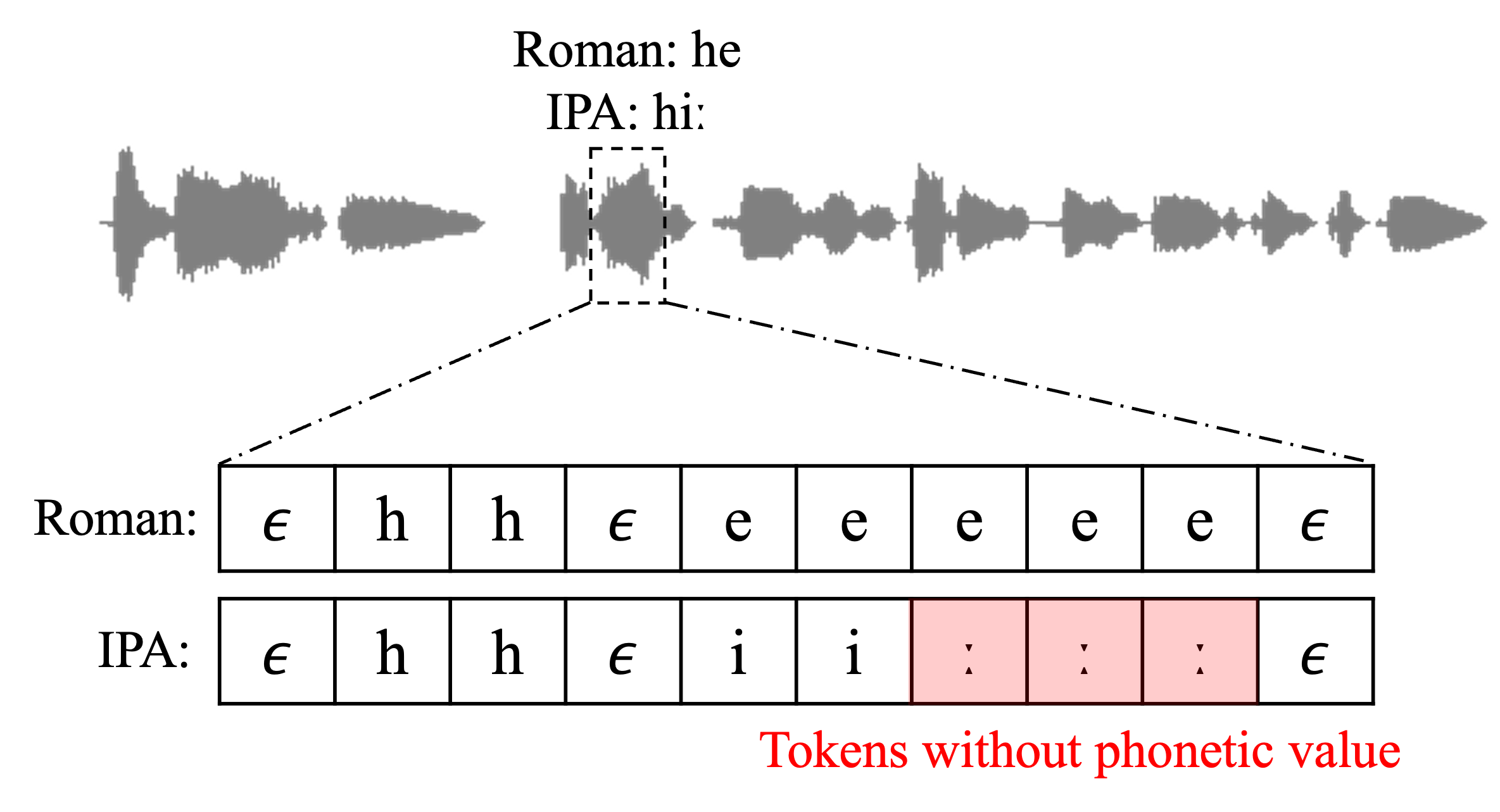}
    \caption{Problems derived in single-token IPA recognition. Diacritic `:' indicates phoneme length, which has no explicit phonetic value. $\epsilon$ denotes a blank token in CTC.}
    \label{fig:problems}
\end{figure}

The first method for orthography unification is to use the international phonetic alphabet (IPA).
IPA is a phonetic notation system that includes four elements: consonants, vowels, diacritics, and suprasegmentals. 
IPA can precisely transcribe pronunciations in a consistent format with a combination of the four elements.
There are challenges with the IPA, especially in vocabulary mapping, and one possible solution is to treat the combination of elements as a single token (e.g., ts, dz, etc.).
However, due to the vast diversity of possible combinations makes this approach difficult to implement.
Conversely, treating each IPA character as a distinct token introduces another issue: characters without phonetic value must be mapped to specific frames as shown in Fig~\ref{fig:problems}.
Since diacritics and suprasegmentals provide detailed information about pronunciation (e.g., length, tone, and stress) but do not carry specific phonetic values, mapping them to distinct frames can introduce confusion during the training process. 

\begin{table*}[h!]
\centering
\small
\begin{tabular}{l||l}
\toprule
\textbf{Zero-Shot}                                                                  & Transcribe following Romanized sentence into a \{lang\} sentence: \{roman\}.                                                                                                                                                                                                 \\ \midrule
\textbf{Few-Shot}                                                                   & \begin{tabular}[l]{@{}l@{}}Here are some examples of transcribing a Romanized sentence into a \{lang\} sentence: \{shots\}.\\Considering the examples above, transcribe the following Romanized sentence into a \{lang\} sentence: \{roman\}.\end{tabular}                     \\ \midrule
\textbf{Zero-Shot CoT}                                                              & Transcribe the following Romanized sentence into a \{lang\} sentence. Think step by step: \{roman\}.                                                                                                                                                                            \\ \midrule
\textbf{\begin{tabular}[l]{@{}l@{}}Few-Shot +\\ Zero-Shot CoT\end{tabular} }        & \begin{tabular}[l]{@{}l@{}}Here are some examples of transcribing a Romanized sentence into a \{lang\} sentence: \{shots\}.\\Considering the examples above, transcribe the following Romanized sentence into a \{lang\} sentence. \\Think step by step: \{roman\}.\end{tabular} \\ \midrule

\textbf{\multirow{2}{*}{\begin{tabular}[l]{@{}l@{}}Prompt\\Chaining\end{tabular}}} & \begin{tabular}[l]{@{}l@{}}Transcribe the following Romanized sentence into a \{lang\} sentence, based on its pronunciation: \{roman\}.\end{tabular}\\ \cmidrule{2-2} & \begin{tabular}[l]{@{}l@{}}Correct the typographical and spacing errors in the following \{lang\} sentence: \{pred\}.\end{tabular}     \\ \bottomrule                                 
\end{tabular}%
\caption{Specific format of the prompt. \textit{roman} refers to the predicted Romanized transcription, \textit{shots} indicates generated examples sampled from the training data, \textit{pred} denotes output from first prompt and \textit{lang} indicates the name of the language.}
\label{prompt_example_tab}
\end{table*}
\subsubsection{Romanization.} Romanization is an alternative method for orthography unification which involves converting text from various writing systems into Latin script.
While Romanization does not preserve phonetic features as precisely as the IPA, it generally retains phonetic information.
Additionally, Romanization offers several advantages over the IPA.
Romanization standardizes diverse writing systems using the Latin alphabet, which is already employed by the majority of languages.
In contrast, IPA requires a specific set of rules for converting the orthography of each language into its IPA representation.
Thus, Romanization is more efficient as it only requires conversion for languages that do not use the Latin alphabet.
Furthermore, Romanization is advantageous for LLMs, as a large portion of their training data consists of Latin characters.
Given these benefits, we adopt Romanization as a method for orthography unification. 

\subsubsection{Universal Transcription Generator.} 
Since our goal is to generate a universal transcription with unified orthography, our first approach was leveraging a wav2vec2.0-phoneme~\cite{xu2021simpleeffectivezeroshotcrosslingual}.
However, we found that directly passing phoneme tokens to the universal converter is suboptimal for transliteration, as it generates phoneme-level tokens without accounting for spacing. 
To address this, we shifted our focus to developing a universal transcription generator that produces character-level tokens while incorporating spacing information.
In this context, Romanization provides a universal character-level orthographic representation, effectively reducing the vocabulary size to around 30 tokens compared to IPA.
Since Romanization aligns with the common phonetic features preserved across languages, we are also confident in the proposed method's strong generalization ability for languages not explicitly included in the training data.
We selected wav2vec2.0-XLS-R~\cite{babu2021xlsrselfsupervisedcrosslingualspeech} with 1 billion parameters, an SSL model pre-trained on 128 languages, as the audio encoder to leverage the advantages of pre-training on a diverse set of languages.
We then attach a classification layer on top and fine-tune both the audio encoder and the classification layer with speech and Romanized transcription pairs to generate universal transcriptions.

\subsection{Language-Specific Transliteration}
The next step is to revert the universal transcription, which retains phonetic features, back to its original language-specific form. 
Since this process involves a text-to-text transformation, we approach it as a transliteration task.
Consequently, we focused on the versatility of LLMs which excel in multilingual and multitask benchmarks due to extensive training on diverse text data.
Therefore, we aim to utilize LLMs as universal converters to transform Romanized transcriptions into language-specific ones.

\subsubsection{Prompt Generation.}
While LLMs have brought a tectonic shift to the NLP domain, additional techniques are still needed to fully harness their emergent abilities.
In this context, prompt engineering has emerged as a field focused on crafting and refining prompts to effectively utilize LLMs across diverse applications and research areas.
To maximize the performance of the inversion process, in the ablation study, we empirically investigated various prompt types: zero-shot, few-shot, zero-shot chain-of-thought (CoT), and prompt chaining, to determine which is the most appropriate for this task. 

\subsubsection{Universal Converter.}
We transliterate the unified Romanized transcription by leveraging LLM's multilingual and multitask language understanding ability without finetuning.
Since our approach does not require any special fine-tuning, the universal converter can be replaced with any superior LLMs, potentially improving the performance of our proposed pipeline in line with the rapidly advancing capabilities of LLMs.
For this paper implementation, we utilize LLaMA3-8B, 4-bit quantized LLaMA3-70B~\cite{touvron2023llamaopenefficientfoundation}, and GPT-4o-mini~\cite{openai2024gpt4technicalreport} as the universal converter.

\section{Experiments}
\subsection{Dataset}
\subsubsection{FLEURS.} FLEURS~\cite{conneau2022fleursfewshotlearningevaluation} is a multilingual speech corpus encompassing 102 languages. It provides a relatively small amount of data per language (approximately 12 hours) while ensuring an unbiased distribution of data across the languages.
Given our focus on demonstrating effective multilingual ASR with minimal data, we utilize the FLEURS and its official splits for experiments.
\subsubsection{CommonVoice.} CommonVoice~\cite{ardila-etal-2020-common} is a multilingual speech dataset crowdsourced from speakers of various languages. For unseen languages, we leverage the official test split of 25 languages from CommonVoice 17.0, which offers sufficient samples for evaluation.

\subsection{Data Preprocessing}
We initially applied NFKC normalization and lowercase transformation to the text transcriptions. 
Subsequently, we excluded samples containing parentheses or numbers from the dataset for the following reasons:
parentheses and digits in transcriptions introduced ambiguity, as some enclosed phrases were pronounced while others were not, and digits had one-to-many pronunciation mappings across languages (e.g. `1' can be pronounced as `one', `eins', `uno', `yi', etc.).
Finally, we utilized the Python library \textit{Uroman}~\cite{hermjakob-etal-2018-box} to obtain Romanized transcription and \textit{Phonemizer}~\cite{Bernard2021} for IPA transcription.
For Japanese, we employed \textit{Pykakasi}~\cite{takahashi1992kakasi} due to the limitation of \textit{Uroman}, which treats Japanese kanji as Chinese characters.
Following these preprocessing steps, we obtained approximately 6 to 8 hours of speech-transcription paired data per language on average.

\begin{table*}[t]
\centering
\small
\begin{tabular}{l||ccccccc|cccc}
\toprule
\multirow{2}{*}{Model}                                                                    & \multicolumn{7}{c|}{Seen (PER / CER $\downarrow$)}                                   & \multicolumn{4}{c}{Unseen (PER / CER $\downarrow$)}                    \\ %\cline{2-12} 
                                                                                          & de & nl  & fr  & es  & it  & \multicolumn{1}{c|}{pt}  & avg  & ia    & eo    & \multicolumn{1}{c|}{eu}    & avg \\ \midrule \midrule
\multirow{2}{*}{\begin{tabular}[c]{@{}l@{}}wav2vec2.0-phoneme\textsuperscript{\dag}\\ + n-gram LM\textsuperscript{\dag}\end{tabular}} & 23.8 & 38.0 & 31.0 & 28.7 & 33.5 & \multicolumn{1}{c|}{45.0} & 33.0 & 10.7 & - & \multicolumn{1}{c|}{20.8} & 31.4   \\
                                                                                          & 14.8 & 26.0  & 26.4  & 12.3    & 21.7    & \multicolumn{1}{c|}{36.5}    & 22.9    & \textbf{6.1}    & -    & \multicolumn{1}{c|}{\textbf{13.7}}    & \textbf{22.2}   \\ \midrule
IPA generator (\shortname)                                                                      & 10.2  & 10.5 & \textbf{9.6} & \textbf{4.2} & 4.6 & \multicolumn{1}{c|}{10.9} & 14.4 & 29.0 & 32.0 & \multicolumn{1}{c|}{36.6} & 35.1  \\ 
\textbf{Roman generator (\shortname)}                                                                    & \textbf{7.3}  & \textbf{9.6} & 12.9 & 4.4 & \textbf{3.6} & \multicolumn{1}{c|}{\textbf{7.2}} & \textbf{11.3} & 14.0 & \textbf{20.8} & \multicolumn{1}{c|}{30.3} & 32.3  \\ \bottomrule
\end{tabular}%
\caption{
Comparison between two orthography unification methods. We report PER and CER for seen and unseen languages. 
Average values are calculated over all 102 seen and 25 unseen languages, respectively. For a fair comparison, all the model sizes are set to 300 million. $\dagger$ denotes results measured by PER, which does not allow for a strict comparison with other results. The language codes and their corresponding language names are provided in the appendix.}
\label{orthography_comp_tab}
\end{table*}
\begin{table}[]
\centering
\small
\begin{tabular}{c|ccc}
\toprule
Model    & \begin{tabular}[c]{@{}c@{}}Data (h)\end{tabular}& \begin{tabular}[c]{@{}c@{}}Universal\end{tabular}  & \begin{tabular}[c]{@{}c@{}}Zero-Shot\end{tabular} \\ 
\midrule\midrule
Whisper & 680k & O &  X \\
MMS-1162   & 122k    & X & X \\
\textbf{\shortname}  &  \textbf{0.6k} & \textbf{O} & \textbf{O} \\ \bottomrule 
\end{tabular}%
\caption{Comparison of previous multi-lingual ASR models with the proposed pipeline. \textit{Data} denotes the total amount of training dataset, \textit{Universal} indicates that no language-specific module is needed, and \textit{Zero-Shot} denotes whether inference on unseen languages is feasible.}
\label{brief_comp_tab}
\end{table}
\subsection{Training Detail}
We performed fine-tuning on all layers except the feature extractor for 3,000 steps with a CTC loss and a batch size of 128.
We bypassed the two-stage fine-tuning pipeline from prior studies~\cite{xu2021simpleeffectivezeroshotcrosslingual, JMLR:v25:23-1318} because our distinct methodology, which used a smaller dataset, caused the divided fine-tuning approach to result in premature convergence and instability.
For hyperparameters, we employed the default AdamW optimizer~\cite{kingma2017adammethodstochasticoptimization, loshchilov2018decoupled} with a tri-stage learning rate scheduler. The warm-up, hold, and decay phases were configured to 10\%, 60\%, and 30\% of the total training steps, respectively.
We then performed a series of experiments to determine the optimal learning rate schedule within the range of 5e-6 to 5e-4.
Finally, the entire training pipeline was conducted on two RTX-3090 GPUs with 24GB of VRAM each, and we leveraged gradient accumulation techniques to address memory issues.

\subsection{Inference Detail}
\subsubsection{Universal Transcription Generator.} 
We leveraged a beam search decoder from flashlight~\cite{kahn2022flashlight} with a beam size of 100. 
No additional lexicons or LMs were utilized in the decoding pipeline to maintain a universal pipeline without relying on language-specific elements.

\subsubsection{Prompting Strategy.}
For the prompting strategy, we utilized language information and a subset of the training data to construct our hypothesis prompt. The specific format of the prompt employed is detailed in Table~\ref{prompt_example_tab}.
In zero-shot prompting, the universal converter automatically transforms Romanized transcriptions into language-specific ones using only the Romanized transcriptions and language information. We employed zero-shot prompting to evaluate the performance of the LLM with minimal input.
Few-shot prompting~\cite{NEURIPS2020_1457c0d6} involves providing examples to help the model generate responses to subsequent instances. 
We hypothesized that this approach would be particularly effective for low-resource or unseen languages by inducing in-context learning. 
Specifically, we randomly sampled five Romanized transcription and target transcription pairs for each few-shot example.
Zero-shot CoT prompting~\cite{NEURIPS2022_8bb0d291} is a technique that supports complex reasoning by inducing the decomposition of intricate tasks into detailed steps. 
Specifically, we appended the phrase ``Let's think step by step'' to the input prompt to encourage the reasoning of the model.
Prompt chaining employs a sequence of prompts, with each prompt building upon the output of the previous one, to manage complex multi-step tasks.
In this aspect, we concentrated on the decomposable process of converting predicted Romanized transcriptions into language-specific transcriptions through \textit{(i) reverse-Romanization} and \textit{(ii) error correction}.
We considered that errors in Romanized transcriptions could propagate during transliteration to language-specific ones, potentially reducing system performance.

\begin{table}[!t]
\centering
\small
\begin{tabular}{c|ccc}
\toprule
                        & Universal Converter       & CER $\downarrow$  & WER $\downarrow$  \\ \midrule\midrule
\multirow{3}{*}{Seen}   & LLaMA-8B    & 26.6 & 46.7 \\
                        & LLaMA-70B   & 15.5 & 35.3 \\
                        & \textbf{GPT-4o-mini} & \textbf{7.5}  & \textbf{18.1} \\ \midrule
\multirow{3}{*}{Unseen} & LLaMA-8B    & 33.0 & 50.2 \\
                        & LLaMA-70B   & 27.2 & 58.9 \\
                        & \textbf{GPT-4o-mini} & \textbf{15.8} & \textbf{38.3} \\ \bottomrule
\end{tabular}%
\caption{
Upper bound performance of the universal converter. 
This upper bound is assessed by feeding ground truth Romanized transcriptions into the universal converter with zero-shot prompting.}
\label{upper_bound_tab}
\end{table}
\subsubsection{Universal Converter.}
Finally, we required the output of the universal converter to conform to a specific format.
We instruct the model to enclose the output within three backticks (e.g., \texttt{```}), which allows us to isolate and sort only the language-specific transcription from the output of the model.
We set the temperature value to 0.0 for all LLMs to obtain deterministic results.
\begin{table*}[]
\centering
\small
\begin{tabular}{c|c||rrr|rrr|rrr}
\toprule
\multirow{2}{*}{Resource} & \multirow{2}{*}{Lang.} & \multicolumn{3}{c|}{Whisper-large-v3}            & \multicolumn{3}{c|}{MMS-1162}                    & \multicolumn{3}{c}{\shortname}   \\ 
                          &                           &Data (h) &CER $\downarrow$  &WER $\downarrow$  &Data (h) &CER $\downarrow$ &WER $\downarrow$ &Data (h) &CER $\downarrow$  &WER $\downarrow$ \\ \midrule\midrule
\multirow{3}{*}{High}     & es                   & 11000            & 1.2  & 3.1  &        2969          &  1.6   &  5.8   & \textbf{6.1}              & 2.8  & 7.3  \\
                          & it                   & 2585             & 0.5  & 1.6  &     1566             &  1.2   &  5.2   & \textbf{6.8}              & 2.0  & 5.2  \\
                          & id                & 1014             & 1.4  & 5.7  &      71            &  2.9   &  14.2   & \textbf{6.8}              & 4.2  & 11.5 \\ \midrule
\multirow{3}{*}{Middle}   & ta                     & 136              & 18.3 & 26.7 &         265         &  11.0   &  41.5   & \textbf{6.3}              & 19.5 & 31.9 \\
                          & ur                      & 104              & 30.9 & 65.0 &      57            &  9.0   &  29.0   & \textbf{4.9}              & 14.9 & 31.9 \\
                          & sk                    & 90               & 2.9  & 8.7  &      301            &  2.2   &  8.8   & \textbf{4.5}              & 3.8  & 10.2 \\ \midrule
\multirow{3}{*}{Low}      & mk                & 16               & 10.3 & 26.3 &       45           &   1.5  &  8.1   & \textbf{5.1}              & 5.5  & 17.2 \\
                          & hi                     & 12               & 35.9 & 43.3 &      57            &  5.8  &   19.6  & \textbf{5.0}              & 8.2  & 15.0 \\ 
                          & kk                   & 12               & 8.5  & 35.1 &         46         &  2.8   &  15.2   & \textbf{8.1}              & 6.7  & 22.9 \\\midrule\midrule
\multicolumn{2}{c||}{Average}                   & -                & 23.9 & 42.9 &         -         &  \textbf{7.8}   &  \textbf{28.8}   & -                & \textbf{14.8} & \textbf{33.2} \\ \bottomrule
\end{tabular}%
\caption{
Comparison results with the baseline models.
Average CER and WER have reported over 82 languages from FLEURS that are covered by Whisper, MMS, and our method. 
The classification of the amount of resources is based on the volume of training data used by Whisper. 
We utilized MMS which encompasses 1162 languages, trained on a combined dataset from MMS-lab, FLEURS, CommonVoice, Voxpopuli~\cite{wang-etal-2021-voxpopuli}, and MLS~\cite{Pratap_2020}.} 
\label{baseline_comparison_tab}
\end{table*}
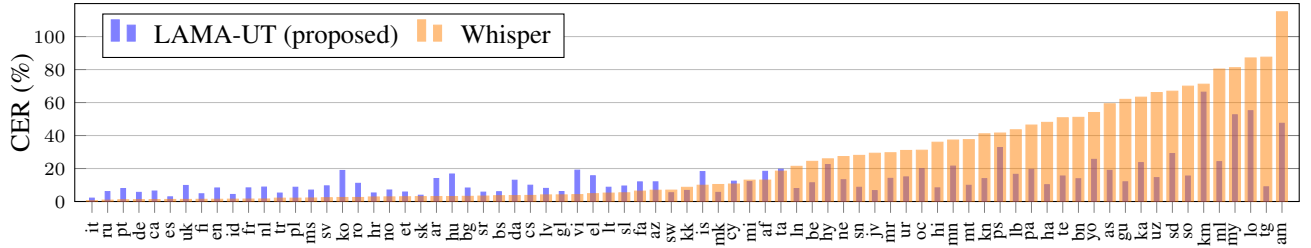
\begin{figure*}[ht]
    \centering
    \begin{tikzpicture}
        \begin{axis}[
            width=\textwidth, 
            height=4.2cm, 
            ylabel={CER (\%)},
            ylabel style={yshift=-7pt},
            xtick={1,2,...,77}, 
            xticklabels={it, ru, pt, de, ca, es, uk, fi, en, id, fr, nl, tr, pl, ms, sv, ko, ro, hr, no, et, sk, ar, hu, bg, sr, bs, da, cs, lv, gl, vi, el, lt, sl, fa, az, sw, kk, is, mk, cy, mi, af, ta, ln, be, hy, ne, sn, jv, mr, ur, oc, hi, mn, mt, kn, ps, lb, pa, ha, te, bn, yo, as, gu, ka, uz, sd, so, km, ml, my, lo, tg, am
            },
            x tick label style={rotate=90, font=\scriptsize},
            xtick pos=bottom,
            grid=major, 
            xmajorgrids=false,
            legend style={at={(0.4,0.95)}, column sep=5pt},%, anchor=south east},
            % legend cell align={left},
            % legend style={at={(0.5, 1)},
            legend cell align={left},
            legend columns=2,
            ymin=0, ymax=120., 
            ytick={0, 20, 40, 60, 80, 100},
            y tick label style={font=\scriptsize},
            xmin=-0.1, xmax=78.1,
            ybar,
            ]
        
        \addplot+[style={blue, opacity=0.5, bar width=0.3, xshift=0.065cm}] 
            coordinates {(1, 2.0434) (2, 5.9955) (3, 7.8678) (4, 5.5343) (5, 6.3371) (6, 2.8268) (7, 9.7158) (8, 4.6651) (9, 8.1988) (10, 4.2642) (11, 8.2475) (12, 8.7417) (13, 5.1053) (14, 8.641) (15, 6.9177) (16, 9.4787) (17, 18.8081) (18, 11.0193) (19, 5.1653) (20, 6.9627) (21, 5.7385) (22, 3.802) (23, 13.936) (24, 16.6712) (25, 8.165) (26, 5.6728) (27, 5.934) (28, 12.8908) (29, 9.879) (30, 7.9353) (31, 6.0453) (32, 18.996) (33, 15.6008) (34, 8.6398) (35, 9.3782) (36, 11.8909) (37, 11.9132) (38, 5.297) (39, 6.7) (40, 18.1172) (41, 5.5156) (42, 12.3207) (43, 12.0296) (44, 18.294) (45, 19.578) (46, 7.8321) (47, 11.334) (48, 22.4033) (49, 13.1836) (50, 8.6191) (51, 6.5947) (52, 14.0207) (53, 14.906) (54, 19.8681) (55, 8.262) (56, 21.4495) (57, 9.8208) (58, 13.8512) (59, 32.7213) (60, 16.4643) (61, 19.5122) (62, 10.1785) (63, 15.4543) (64, 13.7857) (65, 25.5422) (66, 18.8821) (67, 11.9323) (68, 23.5731) (69, 14.4503) (70, 29.063) (71, 15.4388) (72, 66.2708) (73, 24.1487) (74, 52.5678) (75, 55.0304) (76, 8.8922) (77, 47.412)
            };
        \addplot+[style={orange, opacity=0.5, bar width=0.7, xshift=-0.108cm}] 
            coordinates {(1, 0.5042) (2, 0.8322) (3, 1.0532) (4, 1.125) (5, 1.1619) (6, 1.2367) (7, 1.262) (8, 1.3227) (9, 1.3455) (10, 1.4227) (11, 1.4741) (12, 1.5291) (13, 2.0229) (14, 2.0259) (15, 2.1292) (16, 2.2921) (17, 2.435) (18, 2.439) (19, 2.7453) (20, 2.7863) (21, 2.8731) (22, 2.95) (23, 2.9691) (24, 3.0123) (25, 3.1288) (26, 3.2842) (27, 3.5021) (28, 3.5785) (29, 3.6759) (30, 3.9959) (31, 4.0121) (32, 4.1578) (33, 4.7334) (34, 5.0884) (35, 5.2598) (36, 6.2866) (37, 6.8326) (38, 6.8898) (39, 8.5976) (40, 9.8542) (41, 10.3047) (42, 10.5821) (43, 12.9372) (44, 12.9688) (45, 18.3992) (46, 21.3185) (47, 24.3413) (48, 25.8559) (49, 27.2432) (50, 27.9583) (51, 29.2573) (52, 29.5538) (53, 30.9432) (54, 31.085) (55, 35.9342) (56, 37.2523) (57, 37.4938) (58, 41.0351) (59, 41.4955) (60, 43.4683) (61, 46.3415) (62, 47.9747) (63, 50.7616) (64, 51.0178) (65, 53.9499) (66, 59.2416) (67, 61.9198) (68, 63.2461) (69, 66.0443) (70, 66.866) (71, 69.925) (72, 71.1143) (73, 80.2781) (74, 81.1047) (75, 87.0318) (76, 87.4952) (77, 115.0166)};
        
        \legend{LAMA-UT (proposed)\qquad, Whisper}
        \end{axis}
    \end{tikzpicture}
    \caption{CER comparison between \shortname~and Whisper}
    \label{fig:histogram}
\end{figure*}

\section{Results}
Table~\ref{brief_comp_tab} shows that \shortname~effectively achieves multilingual ASR with a universal model.
This approach even operates in a zero-shot environment without requiring language-specific modules while utilizing only a minimal amount of training data.
In the subsequent results, we aim to validate the performance of each component within the pipeline.

\subsection{Universal Transcription Generator}
\subsubsection{Comparison to Baseline Model.}
We conducted a performance comparison between our universal transcription generator and the existing baseline, wav2vec2 phoneme~\cite{xu2021simpleeffectivezeroshotcrosslingual}. Our universal transcription generator focuses on generating character-level tokens and is measured using Character Error Rate (CER), while the baseline wav2vec2 phoneme is measured using Phoneme Error Rate (PER). However, since both metrics are fundamentally used for estimating phonetic symbols, this comparison can be considered meaningful.
Following the Table~\ref{orthography_comp_tab}, the results show that the proposed method demonstrated significantly better performance across a broader range of languages compared to existing approaches even not utilizing language-specific modules (e.g. n-gram LM).
Furthermore, our pipeline demonstrated relatively strong transcription capabilities for unseen languages that were not explicitly included in the training data.
In conclusion, transcribing diverse languages based on their pronunciation can produce a universal transcription which is highly effective.

\subsubsection{Orthography Unification Methods.} 
Among the two methods for standardizing orthographic features, Romanization proved to be more effective than IPA. 
Its ability to represent pronunciation across languages while reducing complexity makes it a superior choice for meaningful results.
Romanization balances phonetic accuracy with simplicity, providing better alignment with LLMs and ensuring efficient processing across multilingual ASR tasks.
 However, since these results are only constrained to the first phase of the pipeline, we have constructed the full end-to-end performance comparison between IPA-based and Romanization-based \shortname, and the results are shown in the appendix.

\subsection{Universal Converter Verification}
Despite the effectiveness of orthography unification, the success of the entire pipeline hinges on the proper functioning of the universal converter. Therefore, the most critical aspect to validate before experimentation was whether a frozen LLM could effectively serve as a universal converter.
To validate this objective, we passed ground truth Romanized transcriptions into the frozen universal converter and assessed its performance. This approach not only tests the converter's capability to accurately produce language-specific transcriptions but also serves to evaluate the upper bound performance of the universal converter within the proposed ASR pipeline.
In Table~\ref{upper_bound_tab}, results demonstrated that universal transcription based on pronunciation characteristics can yield significant performance improvements compared to previous works when the universal transcription generator operates ideally.
However, the upper bound performance for unseen languages showed a slight decrease compared to seen languages. 
This decrease is likely because the unseen languages we tested are typically extremely low-resource languages within the training data of the LLM.

\begin{table*}[]
\centering
\small
\begin{tabular}{c||c|cc|cc|cc|cc|cc}
\toprule
\multirow{4}{*}{Model} & \multirow{4}{*}{\begin{tabular}[c]{@{}c@{}}Repetition \\ \enspace Rate (\%) $\downarrow$  \end{tabular}}  & \multicolumn{10}{c}{Prompting Strategy}                                          \\ %cline{3-12}%\cmidrule{3-12}
                        & & \multicolumn{2}{c}{Zero-Shot} & \multicolumn{2}{c}{\begin{tabular}[c]{@{}c@{}}Few-Shot (5)\end{tabular}} & \multicolumn{2}{c}{Zero-Shot CoT} & \multicolumn{2}{c}{\begin{tabular}[c]{@{}c@{}}Few-Shot (5) +\\ Zero-Shot CoT\end{tabular}} & \multicolumn{2}{c}{\begin{tabular}[c]{@{}c@{}}Prompt\\ Chaining\end{tabular}} \\
                       &  & CER $\downarrow$           & WER $\downarrow$           & CER $\downarrow$                                  & WER $\downarrow$                                 & CER $\downarrow$                                 & WER $\downarrow$                                 & CER $\downarrow$                                       & WER $\downarrow$                                      & CER $\downarrow$                                  & WER $\downarrow$                                  \\ \midrule\midrule
 LLaMA-8B                      & 12        &  35.1             & 70.6              &   22.7                                   &                        49.6             &   37.2                                   &  77.4                                    &    22.1                                      &                                       49.8    &              35.9                        &                              70.8         \\
                        LLaMA-70B           & 1          &       24.3        &        53.8       &       17.4                               &        43.8                             &                       25.4               &     54.6                                 &          16.8                                  &                      43.7                     &      26.7                                 &         55.2                              \\
                        \textbf{GPT-4o-mini}      & \textbf{0.2}               &        \textbf{16.6}       &     \textbf{39.3}          &           \textbf{15.3}                           &                      \textbf{37.2}               &                           18.2           &         41.0                           &                     15.7            &          37.9   &      16.9          &               38.7                   \\ \bottomrule     
\end{tabular}%
\caption{Effects of prompting strategy and model type on the universal converter. The repetition rate indicates the proportion of samples with format errors (e.g., no section enclosed in three backticks until the maximum token limit) due to word repetition.}
\label{tab_ablation}
\end{table*}
\begin{table}[]
\centering
\small
\begin{tabular}{l|cccc}
\toprule
Model       &      \begin{tabular}[c]{@{}c@{}}Data (h)\end{tabular} & \begin{tabular}[c]{@{}c@{}}\# Lang.\end{tabular} & \begin{tabular}[c]{@{}c@{}}Universal\end{tabular} & CER $\downarrow$  \\ \midrule
ASR-2K & 2k & 8 & O                                             & 65.5 \\
% \textbf{\shortname} (IPA)         & \textbf{0.6k} & \textbf{102} & \textbf{O}                                                           & \textbf{40.4} \\
\textbf{\shortname} (Roman)        & \textbf{0.6k} & \textbf{102} & \textbf{O}                                                           & \textbf{34.7} \\
\midrule
MMS-ZS & 40k & 1078 & X                                              & 29.2 \\
+ n-gram LM & 40k & 1078 & X                                           & 25.2 \\
 \bottomrule
\end{tabular}%
\caption{
Comparison with previous zero-shot approaches. 
We evaluated transcription quality on 25 unseen languages from the CommonVoice 17.0 dataset. \textit{\# Lang.} denotes the number of languages leveraged in training.}
\label{unseen_comp_tab}
\end{table}
\subsection{Overall Pipeline}
\subsubsection{Seen Languages.}
We leveraged two baseline models for comparison: Whisper and MMS.
In Table~\ref{baseline_comparison_tab}, results demonstrate that the proposed method achieved a relative reduction of 60\% in CER and 30\% in WER compared to Whisper.
Moreover,~\shortname~matches the performance of MMS despite the absence of language-specific adapters, heads, and n-gram LMs.
Notably, the performance improvements were most pronounced for low-resource languages. 
While Whisper exhibited increased error rates for these languages due to limited training data, our method showed substantial performance enhancements with minimal data resources. 
Despite the slight performance degradation in high-resource languages, the improvement observed in low-resource languages is remarkably meaningful.
The full comparison results are presented in Fig~\ref{fig:histogram}.
It is noteworthy that these results were achieved with considerably smaller training data compared to Whisper and MMS.

\subsubsection{Unseen Languages.}
Our main focus was developing a generalized pipeline that demonstrates strong performance with unseen languages.
To validate this objective, we utilized two zero-shot ASR models as baselines: ASR-2K and Zero-Shot MMS (MMS-ZS).
In Table~\ref{unseen_comp_tab}, our method demonstrated a reduction in CER by half while using significantly less training data compared to ASR-2K.
Furthermore, it is noteworthy that our proposed pipeline performs remarkably well even without language-specific modules, demonstrating comparable performance to MMS-ZS which leverages language-specific lexicon and n-gram LM.

\subsection{Ablation Study}
\subsubsection{Prompting Strategy.}
In Table~\ref{tab_ablation}, few-shot prompting showed the highest performance across all models and prompting strategies.
Interestingly, even with zero-shot prompting, the proposed pipeline consistently outperforms Whisper on average in CER and WER, where Whisper records 23.9\% and 42.9\% respectively, as shown in Table~\ref{baseline_comparison_tab}.
On the other hand, the use of sequential reasoning failed to achieve the anticipated improvements. 
Specifically, we observed considerable error propagation when utilizing zero-shot CoT prompts and prompt chaining techniques.
Minor inaccuracies in the Romanization phase were amplified as they were processed by the LLM, leading to transcriptions that deviated in meaning from the intended output.

\subsubsection{Model Size and Training Data.}
From the perspective of model size, using a relatively smaller LLM like LLaMA-8B frequently resulted in issues such as word repetition, which complicated the transcription sorting process. 
Additionally, this model faced challenges with language misprediction, often generating transcriptions in languages other than the intended target language.
This issue was particularly noticeable with low-resource languages such as Arabic. 
With the LLaMA-70B model, while word repetition was less pronounced compared to the LLaMA-8B model, the issue of language misprediction persisted, albeit at a reduced frequency. 
Among the LLMs tested, GPT-4o-mini demonstrated the best performance overall. It outperformed the other models across all prompting strategies, achieving an impressive average CER of 15\% across 102 languages.

\section{Conclusion}
In this paper, we introduced a generalized multilingual ASR pipeline~\shortname, that operates effectively without relying on language-specific modules. 
By utilizing Romanized transcription as a unified representation across languages, we structured the multilingual ASR pipeline into two phases.
Initially, Romanization aligns phonetic and orthographic features, allowing the universal transcription to be effectively generalized across diverse languages and trained efficiently with a smaller dataset. 
Subsequently, we used a frozen LLM to convert the universal transcription into language-specific ones.
This inversion process showed remarkable performance across languages, including those not previously encountered.
Our experiments demonstrated that the proposed method not only maintains performance for high-resource languages but also significantly outperforms existing methods for low-resource languages, all while effectively handling unseen languages.
Furthermore, our approach matched the performance of models that employ language-specific modules, despite not using any such components. 
We anticipate that this research will provide a viable alternative for utilizing LLMs to support universal multilingual ASR systems across a variety of applications.

\appendix
\section{Appendix}
The following appendix sections propose additional experimental results and specifications for \shortname, which can be instrumental in deeper understanding or validating the strengths of the proposed pipeline.
The first section of the appendix presents the specifications of the languages we leveraged in the experiments, given the corresponding ISO-639 language code.
Consequently, the second section demonstrates the transcription performance of \shortname~for each language used in the experiments.
Then, the third section suggests the end-to-end transcription performance comparison between IPA-based \shortname~and Romanization-based \shortname, further to validate the effectiveness of Romanization in orthography unification.
Finally, in the last section, we would like to discuss possible advancements of \shortname~and our further research direction. 

\section{Languages and Language Codes}
\begin{table*}[!t]
\centering
\small
\begin{tabular}{l||l}
\toprule
Seen   & \begin{tabular}[l]{@{}l@{}}Afrikaans (af), Amharic (am), Arabic (ar), Assamese (as), Asturian (ast), Azerbaijani (az), Belarusian (be),\\Bulgarian (bg), Bengali (bn), Bosnian (bs), Catalan (ca), Cebuano (ceb), Sorani-Kurdish (ku), Mandarin Chinese (cmn),\\Czech (cs), Welsh (cy), Danish (da), German (de), Greek (el), English (en), Spanish (es), Estonian (et), Persian (fa),\\Fula (ff), Finnish (fi), Filipino (fil), French (fr), Irish (ga), Galician (gl), Gujarati (gu), Hausa (ha), Hebrew (he),\\Hindi (hi), Croatian (hr),Hungarian (hu), Armenian (hy), Indonesian (id), Igbo (ig), Icelandic (is), Italian (it),\\Japanese (ja), Javanese (jv), Georgian (ka), Kamba (kam),Kabuverdianu (kea), Kazakh (kk), Khmer (km), Kannada (kn),\\Korean (ko), Kyrgyz (ky), Luxembourgish (lb), Ganda (lg), Lingala (ln), Lao (lo), Lithuanian (lt), Luo (luw), Latvian (lv),\\Maori (mi), Macedonian (mk), Malayalam (ml), Mongolian (mn), Marathi (mr), Malay (ms), Maltese (mt), Burmese (my),\\Norwegian (no), Nepali (ne), Dutch (nl), Northern-Sotho (nso), Nyanja (ny), Occitan (oc), Oromo (om), Oriya (or),\\Punjabi (pa), Polish (pl), Pashto (ps), Portuguese (pt), Romanian (ro), Russian (ru), Sindhi (sd), Slovak (sk), Slovenian (sl),\\Shona (sn), Somali (so), Serbian (sr), Swedish (sv), Swahili (sw), Tamil (ta), Telugu (te), Tajik (tg), Thai (th), Turkish (tr),\\Ukrainian (uk), Umbundu (umb), Urdu (ur), Uzbek (uz), Vietnamese (vi), Wolof (wo), Xhosa (xh), Yoruba (yo),\\Cantonese Chinese (yue), Zulu (zu)\end{tabular}  \\ \midrule
Unseen & \begin{tabular}[l]{@{}l@{}}Abkhazian (ab), Albanian (sq), Basaa (bas), Bashkir (ba), Basque (eu), Breton (br), Chuvash (cv), Eastern Mari (mhr),\\Erzya (myv), Esperanto (eo), Guarani (gn), Hakha Chin (cnh), Interlingua (ia), Kinyarwanda (rw), Latgalian (ltg),\\Norwegian Nynorsk (nn), Romansh (rm), Tatar (tt), Toki Pona (tok), Turkmen (tk), Uighur (ug), Upper Sorbian (hsb),\\Western Frisian (fy), Western Mari (mrj), Yakut (sah)\end{tabular} \\ \bottomrule
\end{tabular}%
\caption{Specifications of 102 seen languages and 25 unseen languages for experiments. We primarily reported ISO-639-1 codes and provided ISO-639-3 codes when the former were unavailable. Sursilvan and Vallader dialects are both considered Romansh languages.}
\label{appendix_langcode}
\end{table*}
Table~\ref{appendix_langcode} indicates the languages that were leveraged in the experiments of our manuscript.
All of the seen languages comprise 102 languages in the FLEURS dataset, and they are used in both training and evaluation of \shortname.
Evaluation samples for unseen languages were chosen from the official test split of the CommonVoice 17.0 dataset, which ensured that they possessed an adequate volume of data.

\section{Language-Specific Performance of \shortname}
\begin{table*}[!th]
\centering
\small
\begin{tabular}{l||l|l||l|l||l|l||l|l}
\toprule
& Language          & CER $\downarrow$ & Language     & CER $\downarrow$ & Language          & CER $\downarrow$ & Language        & CER $\downarrow$ \\ \midrule\midrule
\multirow{26}{*}{Seen} & Afrikaans         & 19.3               & Ganda        & 10.6               & Lithuanian        & 8.7                & Shona           & 9.0                \\
                        & Amharic           & 8.7                & Georgian     & 9.4                & Luo               & 7.4 & Sindhi          & 29.7               \\
                        & Arabic            & 8.8  & German       & 8.4                & Luxembourgish     & 13.6 & Slovak          & 5.4                \\
                        & Armenian          & 5.6 & Greek        & 11.5               & Macedonian        & 5.3                & Slovenian       & 9.1                \\
                        & Assamese          & 12.9               & Gujarati     & 8.9                & Malay             & 8.3                & Somali          & 16.8               \\
                        & Asturian          & 9.3                & Hausa        & 9.7  & Malayalam         & 7.7                & Sorani-Kurdish  & 11.8 \\
                        & Azerbaijani       & 10.7               & Hebrew       & 20.5               & Maltese           & 8.4                & Spanish         & 4.7                \\
                        & Belarusian        & 8.0                & Hindi        & 11.6 & Mandarin Chinese  & 6.5                & Swahili         & 6.4                \\
                        & Bengali           & 9.8                & Hungarian    & 14.5 & Maori             & 8.7                & Swedish         & 11.8 \\
                        & Bosnian           & 6.9                & Icelandic    & 18.7               & Marathi           & 11.9 & Tajik           & 5.3                \\
                        & Bulgarian         & 7.6                & Igbo         & 14.0 & Mongolian         & 14.7               & Tamil           & 11.9 \\
                        & Burmese           & 19.4 & Indonesian   & 6.0                & Nepali            & 11.2 & Telugu          & 10.8               \\
                        & Cantonese Chinese & 16.6               & Irish        & 28.4               & Northern-Sotho    & 13.8               & Thai            & 15.9               \\
                        & Catalan           & 7.5                & Italian      & 3.7 & Norwegian         & 9.2                & Turkish         & 6.4                \\
                        & Cebuano           & 7.5                & Japanese     & 33.4               & Nyanja            & 12.4               & Ukrainian       & 11.2 \\
                        & Croatian          & 6.3                & Javanese     & 7.0  & Occitan           & 17.7               & Umbundu         & 9.5                \\
                        & Czech             & 11.0               & Kabuverdianu & 6.8  & Oriya             & 14.3 & Urdu            & 44.0               \\
                        & Danish            & 15.6               & Kamba        & 10.9               & Oromo             & 20.2 & Uzbek           & 16.3               \\
                        & Dutch             & 11.3               & Kannada      & 7.4 & Pashto            & 20.0               & Vietnamese      & 13.0               \\
                        & English           & 13.1 & Kazakh       & 5.1                & Persian           & 6.9                & Welsh           & 12.8               \\
                        & Estonian          & 4.4 & Khmer        & 23.6 & Polish            & 10.9               & Wolof           & 14.9 \\
                        & Filipino          & 4.8                & Korean       & 9.6                & Portuguese        & 9.0                & Xhosa           & 10.4               \\
                        & Finnish           & 4.9                & Kyrgyz       & 5.2                & Punjabi           & 20.3               & Yoruba          & 10.9               \\
                        & French            & 12.3               & Lao          & 17.8               & Romanian          & 11.8 & Zulu            & 9.5                \\
                        & Fula              & 14.7               & Latvian      & 7.0  & Russian           & 9.0                &                 &                    \\
                        & Galician          & 6.7                & Lingala      & 7.7                & Serbian           & 8.9                &                 &                    \\ \midrule
\multirow{7}{*}{Unseen} & Abkhazian         & 45.8 & Eastern Mari & 30.9               & Latgalian         & 28.8 & Upper Sorbian   & 35.3               \\
                        & Albanian          & 34.8               & Erzya        & 29.9               & Norwegian Nynorsk & 27.1               & Western Frisian & 33.8 \\
                        & Basa              & 33.4               & Esperanto    & 22.1               & Romansh           & 31.9 & Western Mari    & 36.7               \\
                        & Bashkir           & 33.5               & Guarani      & 33.5               & Tatar             & 31.8               & Yakut           & 36.3               \\
                        & Basque            & 33.0               & Hakha Chin   & 42.9               & Toki Pona         & 28.7               &                 &                    \\
                        & Breton            & 49.1               & Interlingua  & 17.3 & Turkmen           & 41.3               &                 &                    \\
                        & Chuvash           & 37.4               & Kinyarwanda  & 37.2               & Uighur            & 27.1               &                 &                  \\ \bottomrule
\end{tabular}
\caption{Performance of Romanization-based universal transcription generator in 102 seen and 25 unseen languages. Results are reported in CER, which calculates the character-level difference between ground truth Romanized transcription and predicted Romanized transcription.}
\label{appendix_cer_roman_all}
\end{table*}
\begin{table*}[!th]
\centering
\small
\begin{tabular}{l||l|l||l|l||l|l||l|l}
\toprule
& Language          & CER $\downarrow$ & Language     & CER $\downarrow$ & Language          & CER $\downarrow$ & Language        & CER $\downarrow$ \\ \midrule\midrule
\multirow{26}{*}{Seen} & Afrikaans         & 18.3        & Ganda        & 10.4        & Luo               & 7.3         & Sindhi          & 29.1        \\
& \textbf{Amharic}           & \textbf{47.4}        & Georgian     & 23.6        & Luxembourgish     & 16.5        & Slovak          & 3.8         \\
& Arabic            & 13.9        & German       & 5.5         & Macedonian        & 5.5         & Slovenian       & 9.4         \\
& \textbf{Armenian}          & \textbf{22.4}        & Greek        & 15.6        & Malay             & 6.9         & Somali          & 15.4        \\
& Assamese          & 18.9        & Gujarati     & 11.9        & \textbf{Malayalam}         & \textbf{24.1}        & Sorani-Kurdish  & 26.4        \\
& Asturian          & 9.8         & Hausa        & 10.2        & Maltese           & 9.8         & Spanish         & 2.8         \\
& Azerbaijani       & 11.9        & \textbf{Hebrew}       & \textbf{35.2}        & \textbf{Mandarin Chinese}  & \textbf{36.0}        & Swahili         & 5.3         \\
& Belarusian        & 11.3        & Hindi        & 8.3         & Maori             & 12.0        & Swedish         & 9.5         \\
& Bengali           & 13.8        & Hungarian    & 16.7        & Marathi           & 14.0        & Tajik           & 8.9         \\
& Bosnian           & 5.9         & Icelandic    & 18.1        & Mongolian         & 21.4        & Tamil           & 19.6        \\
& Bulgarian         & 8.2         & Igbo         & 17.9        & Nepali            & 13.2        & Telugu          & 15.5        \\
& \textbf{Burmese}           & \textbf{52.6}        & Indonesian   & 4.3         & Northern-Sotho    & 15.1        & \textbf{Thai}            & \textbf{45.3}        \\
& \textbf{Cantonese Chinese} & \textbf{61.6}        & Irish        & 34.8        & Norwegian         & 7.0         & Turkish         & 5.1         \\
& Catalan           & 6.3         & Italian      & 2.0         & Nyanja            & 12.1        & Ukrainian       & 9.7         \\
& Cebuano           & 7.4         & Javanese     & 6.6         & Occitan           & 19.9        & Umbundu         & 10.5        \\
& Croatian          & 5.2         & Kabuverdianu & 8.3         & Oriya             & 18.3        & Urdu            & 14.9        \\
& Czech             & 9.9         & Kamba        & 14.6        & Oromo             & 19.8        & Uzbek           & 14.5        \\
& Danish            & 12.9        & Kannada      & 13.9        & Pashto            & 32.7        & Vietnamese      & 19.0        \\
& Dutch             & 8.7         & Kazakh       & 6.7         & Persian           & 11.9        & Welsh           & 12.3        \\
& English           & 8.2         & \textbf{Khmer}        & \textbf{66.3}        & Polish            & 8.6         & Wolof           & 18.0        \\
& Estonian          & 5.7         & Korean       & 18.8        & Portuguese        & 7.9         & Xhosa           & 10.5        \\
& Filipino          & 4.4         & Kyrgyz       & 8.8         & Punjabi           & 19.5        & Yoruba          & 25.5        \\
& Finnish           & 4.7         & \textbf{Lao}          & \textbf{55.0}        & Romanian          & 11.0        & Zulu            & 9.6         \\
& French            & 8.2         & Latvian      & 7.9         & Russian           & 6.0         &                 &             \\
& Fula              & 16.6        & Lingala      & 7.8         & Serbian           & 5.7         &                 &             \\
& Galician          & 6.0         & Lithuanian   & 8.6         & Shona             & 8.6         &                 &             \\ \midrule
\multirow{7}{*}{Unseen} & Abkhazian         & 55.9        & Eastern Mari & 35.3        & Latgalian         & 34.7        & Upper Sorbian   & 35.2        \\
& Albanian          & 35.4        & Erzya        & 35.7        & Norwegian Nynorsk & 23.9        & Western Frisian & 29.6        \\
& Basa              & 42.1        & Esperanto    & 19.7        & Romansh           & 30.0        & Western Mari    & 41.4        \\
& Bashkir           & 29.7        & Guarani      & 40.0        & Tatar             & 24.2        & Yakut           & 38.2        \\
& Basque            & 32.8        & Hakha Chin   & 42.8        & Toki Pona         & 29.0        &                 &             \\
& Breton            & 49.9        & Interlingua  & 15.2        & Turkmen           & 43.1        &                 &             \\
& Chuvash           & 44.3        & Kinyarwanda  & 36.7        & Uighur            & 24.5        &                 &             \\ \bottomrule
\end{tabular}
\caption{Performance of LAMA-UT in 101 seen and 25 unseen languages. We leveraged a Romanization-based universal transcription generator and GPT-4o-mini model with a few-shot prompting as a universal converter. All of the results are reported in CER, which calculates the character-level difference between ground truth language-specific transcription and predicted language-specific transcription. Japanese was excluded from the results due to the ambiguity in evaluation criteria arising from its mixed use of Hiragana, Katakana, and Kanji.}
\label{appendix_performance_every_all}
\vspace{10pt}
\end{table*}
In this section, we present a detailed performance analysis for each language, comparing transcription performance from the universal transcription generation phase to that after passing through the universal converter. 
Additionally, we analyze these results further to explore the correlation between orthographic characteristics and transcription performance.
First of all, Table ~\ref{appendix_cer_roman_all} suggests the language-specific transcription performance of the Romanization-based universal transcription generator, and Table~\ref{appendix_performance_every_all} presents the language-specific end-to-end (E2E) language-specific transcription performance of \shortname.
In the case of the universal transcription generator, CERs were consistently similar across most languages with minimal variation. However, in the end-to-end pipeline with language-specific transliteration, notably higher error rates were observed for certain languages, and the languages highlighted in bold represent the top 10 languages with the most significant increases in error rate observed in the end-to-end pipeline compared to the CER values during the universal transcription phase.
Interestingly, the 10 languages that exhibited the most significant performance degradation after passing through the universal converter were all non-Latin script languages, with the majority being languages that do not employ spacing in their orthography.
This suggests that when a pre-trained LLM is utilized as a universal converter, languages with non-Latin orthography which inherently exhibit different structural characteristics compared to Latin-based languages, are more prone to error propagation.

\section{E2E Comparison on Unification Methods}
\begin{table*}[th!]
\centering
\begin{tabular}{c||c||cc|cc||cc|cc}
\toprule
\multirow{3}{*}{}       & \multirow{3}{*}{\begin{tabular}[c]{@{}c@{}}Universal \\ Converter  \end{tabular}} & \multicolumn{4}{c}{IPA}                                        & \multicolumn{4}{c}{Roman}                                      \\
                        &                                      & \multicolumn{2}{c}{Upper Bound} & \multicolumn{2}{c}{Few Shot} & \multicolumn{2}{c}{Upper Bound} & \multicolumn{2}{c}{Few Shot} \\
                        &                                      & CER $\downarrow$            & WER $\downarrow$            & CER $\downarrow$           & WER $\downarrow$          & CER $\downarrow$            & WER $\downarrow$            & CER $\downarrow$           & WER $\downarrow$          \\ \midrule
\multirow{3}{*}{Seen}   & LLaMA-8B                             &               51.4 &               82.4 &              56.5 &             80.8 &               33.0 &               44.4 &              38.1 &             64.7 \\
                        & LLaMA-70B                            &               40.0 &               64.6 &              33.8 &             67.8 &               15.6 &               34.6 &              23.1 &             56.2 \\
                        & GPT-4o-mini                          &               27.3 &               49.2 &              33.0 &             70.6 &               10.0 &               28.6 &              19.9 &             49.4 \\ \midrule
\multirow{3}{*}{Unseen} & LLaMA-8B                             &               52.0 &               93.0 &              43.4 &             93.3 &               32.7 &               60.2 &              41.6 &             94.4 \\
                        & LLaMA-70B                            &               42.5 &               89.0 &              38.1 &             94.6 &               17.0 &               57.5 &              34.7 &             93.9 \\
                        & GPT-4o-mini                          &               40.4 &               78.2 &              35.2 &             84.1 &               12.3 &               42.1 &              29.7 &            83.0 \\ \bottomrule
\end{tabular}
\caption{An end-to-end (E2E) comparison between IPA and Romanization. Since \textit{Phonemizer} (IPA transliterator) supports a smaller amount of language compared to \textit{Uroman} (Romanization transliterator), all of the metrics were calculated and averaged within the common languages which are both available in the IPA-based and the Romanization-based version of the universal transcription generator.}
\label{appendix_e2e}
\end{table*}
In this section, we would like to further clarify the effectiveness of utilizing Romanization as an intermediate representation.
Since the experimental results from Table~\ref{orthography_comp_tab} were limited to the universal transcription generation phase, we
supplement the previous results with the full performance comparison between IPA-based and Romanization-based end-to-end architecture of \shortname, both in upper bound assessment and transcription performance.
Table~\ref{appendix_e2e} proposes the end-to-end performance comparison between IPA-based \shortname~and Romanization-based \shortname.
The only difference between the two models is the orthography unification method utilized in the universal transcription generator. (e.g., the IPA-based model utilizes IPA as an intermediate representation, while the Romanization-based one leverages Romanized transcription as an intermediate representation.
In the results, Romanization-based \shortname~consistently outperforms IPA-based \shortname~as a substantial difference in both CER and WER.
These results strongly demonstrate the superiority of Romanization over IPA when leveraging LLMs as a universal converter, since LLMs are primarily trained with Latin scripts and optimized tokenization strategy for them.
Furthermore, there are some inconsistencies in the results of the IPA-based \shortname, where passing predicted IPA transcriptions along with a few examples to the universal converter yielded better performance than passing ground truth IPA transcriptions without examples.
This is presumably because the LLMs did not frequently encounter IPA-driven tokens during its training process.

\section{Discussion and Future Work}
\shortname~showed comparable or better performance compared to previous works even without language-specific modules (e.g., adapters, lexicons, n-gram LMs), while achieving efficient training with a significantly reduced dataset size.
However, there is still room for further improvement in both the universal transcription generator and the universal converter.
First, the transcription performance of the universal transcription generator can be improved. For instance, the universal transcription generator of \shortname~can leverage additional linguistic information (e.g., embedding from a pre-trained language classifier) to further enhance the transcription quality of the first phase.
Secondly, our pipeline shows relatively lower performance for languages with distinct linguistic structures, like Korean, and those with additional features (e.g., tones), such as Chinese, in the language-specific transliteration phase.
Since our universal converter is replaceable, this issue will naturally be resolved in line with the development of LLMs.
Finally, the utilization of prompt learning techniques~\cite{li2021prefixtuningoptimizingcontinuousprompts, liu2022ptuningv2prompttuning, gu-etal-2022-ppt} might improve transliteration performance.
We plan to address these aspects in future research.
\vspace{30pt}

\bibliography{aaai25}

\begin{thebibliography}{52}
\providecommand{\natexlab}[1]{#1}

\bibitem[{Ardila et~al.(2020)Ardila, Branson, Davis, Kohler, Meyer, Henretty, Morais, Saunders, Tyers, and Weber}]{ardila-etal-2020-common}
Ardila, R.; Branson, M.; Davis, K.; Kohler, M.; Meyer, J.; Henretty, M.; Morais, R.; Saunders, L.; Tyers, F.; and Weber, G. 2020.
\newblock Common Voice: A Massively-Multilingual Speech Corpus.
\newblock In \emph{Proceedings of the Twelfth Language Resources and Evaluation Conference}, 4218--4222. European Language Resources Association.

\bibitem[{Babu et~al.(2021)Babu, Wang, Tjandra, Lakhotia, Xu, Goyal, Singh, von Platen, Saraf, Pino, Baevski, Conneau, and Auli}]{babu2021xlsrselfsupervisedcrosslingualspeech}
Babu, A.; Wang, C.; Tjandra, A.; Lakhotia, K.; Xu, Q.; Goyal, N.; Singh, K.; von Platen, P.; Saraf, Y.; Pino, J.; Baevski, A.; Conneau, A.; and Auli, M. 2021.
\newblock XLS-R: Self-supervised Cross-lingual Speech Representation Learning at Scale.
\newblock arXiv:2111.09296.

\bibitem[{Baevski et~al.(2020)Baevski, Zhou, Mohamed, and Auli}]{NEURIPS2020_92d1e1eb}
Baevski, A.; Zhou, Y.; Mohamed, A.; and Auli, M. 2020.
\newblock wav2vec 2.0: A Framework for Self-Supervised Learning of Speech Representations.
\newblock In \emph{Advances in Neural Information Processing Systems}, volume~33, 12449--12460. Curran Associates, Inc.

\bibitem[{Bernard and Titeux(2021)}]{Bernard2021}
Bernard, M.; and Titeux, H. 2021.
\newblock Phonemizer: Text to Phones Transcription for Multiple Languages in Python.
\newblock \emph{Journal of Open Source Software}, 6(68): 3958.

\bibitem[{Brown et~al.(2020)Brown, Mann, Ryder, Subbiah, Kaplan, Dhariwal, Neelakantan, Shyam, Sastry, Askell, Agarwal, Herbert-Voss, Krueger, Henighan, Child, Ramesh, Ziegler, Wu, Winter, Hesse, Chen, Sigler, Litwin, Gray, Chess, Clark, Berner, McCandlish, Radford, Sutskever, and Amodei}]{NEURIPS2020_1457c0d6}
Brown, T.; Mann, B.; Ryder, N.; Subbiah, M.; Kaplan, J.~D.; Dhariwal, P.; Neelakantan, A.; Shyam, P.; Sastry, G.; Askell, A.; Agarwal, S.; Herbert-Voss, A.; Krueger, G.; Henighan, T.; Child, R.; Ramesh, A.; Ziegler, D.; Wu, J.; Winter, C.; Hesse, C.; Chen, M.; Sigler, E.; Litwin, M.; Gray, S.; Chess, B.; Clark, J.; Berner, C.; McCandlish, S.; Radford, A.; Sutskever, I.; and Amodei, D. 2020.
\newblock Language Models are Few-Shot Learners.
\newblock In \emph{Advances in Neural Information Processing Systems}, volume~33, 1877--1901. Curran Associates, Inc.

\bibitem[{Chan et~al.(2016)Chan, Jaitly, Le, and Vinyals}]{chan2016listen}
Chan, W.; Jaitly, N.; Le, Q.; and Vinyals, O. 2016.
\newblock Listen, attend and spell: A neural network for large vocabulary conversational speech recognition.
\newblock In \emph{2016 IEEE international conference on acoustics, speech and signal processing (ICASSP)}, 4960--4964. IEEE.

\bibitem[{Chen et~al.(2022)Chen, Wang, Chen, Wu, Liu, Chen, Li, Kanda, Yoshioka, Xiao et~al.}]{chen2022wavlm}
Chen, S.; Wang, C.; Chen, Z.; Wu, Y.; Liu, S.; Chen, Z.; Li, J.; Kanda, N.; Yoshioka, T.; Xiao, X.; et~al. 2022.
\newblock Wavlm: Large-scale self-supervised pre-training for full stack speech processing.
\newblock \emph{IEEE Journal of Selected Topics in Signal Processing}, 16(6): 1505--1518.

\bibitem[{Chorowski and Jaitly(2016)}]{chorowski2016betterdecodinglanguagemodel}
Chorowski, J.; and Jaitly, N. 2016.
\newblock Towards better decoding and language model integration in sequence to sequence models.
\newblock arXiv:1612.02695.

\bibitem[{Clark, Yallop, and Fletcher(2007)}]{clark2007introduction}
Clark, J.; Yallop, C.; and Fletcher, J. 2007.
\newblock \emph{An Introduction to Phonetics and Phonology}.
\newblock Blackwell Textbooks in Linguistics. Wiley.
\newblock ISBN 9781405130837.

\bibitem[{Conneau et~al.(2020)Conneau, Baevski, Collobert, Mohamed, and Auli}]{conneau2020unsupervisedcrosslingualrepresentationlearning}
Conneau, A.; Baevski, A.; Collobert, R.; Mohamed, A.; and Auli, M. 2020.
\newblock Unsupervised Cross-lingual Representation Learning for Speech Recognition.
\newblock arXiv:2006.13979.

\bibitem[{Conneau et~al.(2022)Conneau, Ma, Khanuja, Zhang, Axelrod, Dalmia, Riesa, Rivera, and Bapna}]{conneau2022fleursfewshotlearningevaluation}
Conneau, A.; Ma, M.; Khanuja, S.; Zhang, Y.; Axelrod, V.; Dalmia, S.; Riesa, J.; Rivera, C.; and Bapna, A. 2022.
\newblock FLEURS: Few-shot Learning Evaluation of Universal Representations of Speech.
\newblock arXiv:2205.12446.

\bibitem[{Du et~al.(2024)Du, Wang, Chen, Chu, Gao, Li, Hu, Zhou, Xu, Ma, Wang, Zheng, Zhou, Yan, and Zhang}]{du2024lauragptlistenattendunderstand}
Du, Z.; Wang, J.; Chen, Q.; Chu, Y.; Gao, Z.; Li, Z.; Hu, K.; Zhou, X.; Xu, J.; Ma, Z.; Wang, W.; Zheng, S.; Zhou, C.; Yan, Z.; and Zhang, S. 2024.
\newblock LauraGPT: Listen, Attend, Understand, and Regenerate Audio with GPT.
\newblock arXiv:2310.04673.

\bibitem[{Fathullah et~al.(2024)Fathullah, Wu, Lakomkin, Jia, Shangguan, Li, Guo, Xiong, Mahadeokar, Kalinli et~al.}]{fathullah2024prompting}
Fathullah, Y.; Wu, C.; Lakomkin, E.; Jia, J.; Shangguan, Y.; Li, K.; Guo, J.; Xiong, W.; Mahadeokar, J.; Kalinli, O.; et~al. 2024.
\newblock Prompting large language models with speech recognition abilities.
\newblock In \emph{ICASSP 2024-2024 IEEE International Conference on Acoustics, Speech and Signal Processing (ICASSP)}, 13351--13355. IEEE.

\bibitem[{Graves(2012)}]{graves2012sequencetransductionrecurrentneural}
Graves, A. 2012.
\newblock Sequence Transduction with Recurrent Neural Networks.
\newblock arXiv:1211.3711.

\bibitem[{Graves et~al.(2006)Graves, Fern{\'a}ndez, Gomez, and Schmidhuber}]{graves2006connectionist}
Graves, A.; Fern{\'a}ndez, S.; Gomez, F.; and Schmidhuber, J. 2006.
\newblock Connectionist temporal classification: labelling unsegmented sequence data with recurrent neural networks.
\newblock In \emph{Proceedings of the 23rd international conference on Machine learning}, 369--376.

\bibitem[{Gu et~al.(2022)Gu, Han, Liu, and Huang}]{gu-etal-2022-ppt}
Gu, Y.; Han, X.; Liu, Z.; and Huang, M. 2022.
\newblock {PPT}: Pre-trained Prompt Tuning for Few-shot Learning.
\newblock In \emph{Proceedings of the 60th Annual Meeting of the Association for Computational Linguistics (Volume 1: Long Papers)}, 8410--8423. Association for Computational Linguistics.

\bibitem[{Gulati et~al.(2020)Gulati, Qin, Chiu, Parmar, Zhang, Yu, Han, Wang, Zhang, Wu, and Pang}]{gulati20_interspeech}
Gulati, A.; Qin, J.; Chiu, C.-C.; Parmar, N.; Zhang, Y.; Yu, J.; Han, W.; Wang, S.; Zhang, Z.; Wu, Y.; and Pang, R. 2020.
\newblock Conformer: Convolution-augmented Transformer for Speech Recognition.
\newblock In \emph{Interspeech 2020}, 5036--5040.

\bibitem[{Hermjakob, May, and Knight(2018)}]{hermjakob-etal-2018-box}
Hermjakob, U.; May, J.; and Knight, K. 2018.
\newblock Out-of-the-box Universal {R}omanization Tool uroman.
\newblock In \emph{Proceedings of {ACL} 2018, System Demonstrations}, 13--18. Association for Computational Linguistics.

\bibitem[{Hsu et~al.(2021)Hsu, Bolte, Tsai, Lakhotia, Salakhutdinov, and Mohamed}]{hsu2021hubert}
Hsu, W.-N.; Bolte, B.; Tsai, Y.-H.~H.; Lakhotia, K.; Salakhutdinov, R.; and Mohamed, A. 2021.
\newblock Hubert: Self-supervised speech representation learning by masked prediction of hidden units.
\newblock \emph{IEEE/ACM transactions on audio, speech, and language processing}, 29: 3451--3460.

\bibitem[{Hu et~al.(2023)Hu, Sainath, Li, Du, Huang, Dai, Zhang, Cabrera, Chen, and Strohman}]{hu2023massively}
Hu, K.; Sainath, T.~N.; Li, B.; Du, N.; Huang, Y.; Dai, A.~M.; Zhang, Y.; Cabrera, R.; Chen, Z.; and Strohman, T. 2023.
\newblock Massively multilingual shallow fusion with large language models.
\newblock In \emph{ICASSP 2023-2023 IEEE International Conference on Acoustics, Speech and Signal Processing (ICASSP)}, 1--5. IEEE.

\bibitem[{Huang et~al.(2024)Huang, Allauzen, Chen, Gupta, Hu, Qin, Zhang, Wang, Chang, and Sainath}]{huang2024multilingual}
Huang, W.~R.; Allauzen, C.; Chen, T.; Gupta, K.; Hu, K.; Qin, J.; Zhang, Y.; Wang, Y.; Chang, S.-Y.; and Sainath, T.~N. 2024.
\newblock Multilingual and fully non-autoregressive asr with large language model fusion: A comprehensive study.
\newblock In \emph{ICASSP 2024-2024 IEEE International Conference on Acoustics, Speech and Signal Processing (ICASSP)}, 13306--13310. IEEE.

\bibitem[{Kahn et~al.(2022)Kahn, Pratap, Likhomanenko, Xu, Hannun, Cai, Tomasello, Lee, Grave, Avidov et~al.}]{kahn2022flashlight}
Kahn, J.~D.; Pratap, V.; Likhomanenko, T.; Xu, Q.; Hannun, A.; Cai, J.; Tomasello, P.; Lee, A.; Grave, E.; Avidov, G.; et~al. 2022.
\newblock Flashlight: Enabling innovation in tools for machine learning.
\newblock In \emph{International Conference on Machine Learning}, 10557--10574. PMLR.

\bibitem[{Kannan et~al.(2018)Kannan, Wu, Nguyen, Sainath, Chen, and Prabhavalkar}]{kannan2018analysis}
Kannan, A.; Wu, Y.; Nguyen, P.; Sainath, T.~N.; Chen, Z.; and Prabhavalkar, R. 2018.
\newblock An analysis of incorporating an external language model into a sequence-to-sequence model.
\newblock In \emph{2018 IEEE International Conference on Acoustics, Speech and Signal Processing (ICASSP)}, 1--5828. IEEE.

\bibitem[{Kingma and Ba(2017)}]{kingma2017adammethodstochasticoptimization}
Kingma, D.~P.; and Ba, J. 2017.
\newblock Adam: A Method for Stochastic Optimization.
\newblock arXiv:1412.6980.

\bibitem[{Kojima et~al.(2022)Kojima, Gu, Reid, Matsuo, and Iwasawa}]{NEURIPS2022_8bb0d291}
Kojima, T.; Gu, S.~S.; Reid, M.; Matsuo, Y.; and Iwasawa, Y. 2022.
\newblock Large Language Models are Zero-Shot Reasoners.
\newblock In \emph{Advances in Neural Information Processing Systems}, volume~35, 22199--22213. Curran Associates, Inc.

\bibitem[{Ladefoged and Maddieson(1996)}]{ladefoged1996sounds}
Ladefoged, P.; and Maddieson, I. 1996.
\newblock \emph{The sounds of the world's languages}, volume 1012.
\newblock Blackwell Oxford.

\bibitem[{Li et~al.(2020)Li, Dalmia, Li, Lee, Littell, Yao, Anastasopoulos, Mortensen, Neubig, Black et~al.}]{li2020universal}
Li, X.; Dalmia, S.; Li, J.; Lee, M.; Littell, P.; Yao, J.; Anastasopoulos, A.; Mortensen, D.~R.; Neubig, G.; Black, A.~W.; et~al. 2020.
\newblock Universal phone recognition with a multilingual allophone system.
\newblock In \emph{ICASSP 2020-2020 IEEE International Conference on Acoustics, Speech and Signal Processing (ICASSP)}, 8249--8253. IEEE.

\bibitem[{Li et~al.(2022{\natexlab{a}})Li, Metze, Mortensen, Black, and Watanabe}]{li22aa_interspeech}
Li, X.; Metze, F.; Mortensen, D.~R.; Black, A.~W.; and Watanabe, S. 2022{\natexlab{a}}.
\newblock {ASR2K: Speech Recognition for Around 2000 Languages without Audio}.
\newblock In \emph{Proc. Interspeech 2022}, 4885--4889.

\bibitem[{Li et~al.(2022{\natexlab{b}})Li, Metze, Mortensen, Watanabe, and Black}]{li2022zero}
Li, X.; Metze, F.; Mortensen, D.~R.; Watanabe, S.; and Black, A.~W. 2022{\natexlab{b}}.
\newblock Zero-shot learning for grapheme to phoneme conversion with language ensemble.
\newblock In \emph{Findings of the Association for Computational Linguistics: ACL 2022}, 2106--2115.

\bibitem[{Li and Liang(2021)}]{li2021prefixtuningoptimizingcontinuousprompts}
Li, X.~L.; and Liang, P. 2021.
\newblock Prefix-Tuning: Optimizing Continuous Prompts for Generation.
\newblock arXiv:2101.00190.

\bibitem[{Li et~al.(2023)Li, Wu, Li, and Liu}]{li2023prompting}
Li, Y.; Wu, Y.; Li, J.; and Liu, S. 2023.
\newblock Prompting large language models for zero-shot domain adaptation in speech recognition.
\newblock In \emph{2023 IEEE Automatic Speech Recognition and Understanding Workshop (ASRU)}, 1--8. IEEE.

\bibitem[{Liu et~al.(2022)Liu, Ji, Fu, Tam, Du, Yang, and Tang}]{liu2022ptuningv2prompttuning}
Liu, X.; Ji, K.; Fu, Y.; Tam, W.~L.; Du, Z.; Yang, Z.; and Tang, J. 2022.
\newblock P-Tuning v2: Prompt Tuning Can Be Comparable to Fine-tuning Universally Across Scales and Tasks.
\newblock arXiv:2110.07602.

\bibitem[{Loshchilov and Hutter(2019)}]{loshchilov2018decoupled}
Loshchilov, I.; and Hutter, F. 2019.
\newblock Decoupled Weight Decay Regularization.
\newblock In \emph{International Conference on Learning Representations}.

\bibitem[{OpenAI(2024)}]{openai2024gpt4technicalreport}
OpenAI. 2024.
\newblock GPT-4 Technical Report.
\newblock arXiv:2303.08774.

\bibitem[{Park et~al.(2020)Park, Zhang, Jia, Han, Chiu, Li, Wu, and Le}]{Park_2020}
Park, D.~S.; Zhang, Y.; Jia, Y.; Han, W.; Chiu, C.-C.; Li, B.; Wu, Y.; and Le, Q.~V. 2020.
\newblock Improved Noisy Student Training for Automatic Speech Recognition.
\newblock In \emph{Interspeech 2020}. ISCA.

\bibitem[{Pratap et~al.(2020{\natexlab{a}})Pratap, Sriram, Tomasello, Hannun, Liptchinsky, Synnaeve, and Collobert}]{pratap2020massivelymultilingualasr50}
Pratap, V.; Sriram, A.; Tomasello, P.; Hannun, A.; Liptchinsky, V.; Synnaeve, G.; and Collobert, R. 2020{\natexlab{a}}.
\newblock Massively Multilingual ASR: 50 Languages, 1 Model, 1 Billion Parameters.
\newblock arXiv:2007.03001.

\bibitem[{Pratap et~al.(2024)Pratap, Tjandra, Shi, Tomasello, Babu, Kundu, Elkahky, Ni, Vyas, Fazel-Zarandi, Baevski, Adi, Zhang, Hsu, Conneau, and Auli}]{JMLR:v25:23-1318}
Pratap, V.; Tjandra, A.; Shi, B.; Tomasello, P.; Babu, A.; Kundu, S.; Elkahky, A.; Ni, Z.; Vyas, A.; Fazel-Zarandi, M.; Baevski, A.; Adi, Y.; Zhang, X.; Hsu, W.-N.; Conneau, A.; and Auli, M. 2024.
\newblock Scaling Speech Technology to 1,000+ Languages.
\newblock \emph{Journal of Machine Learning Research}, 25(97): 1--52.

\bibitem[{Pratap et~al.(2020{\natexlab{b}})Pratap, Xu, Sriram, Synnaeve, and Collobert}]{Pratap_2020}
Pratap, V.; Xu, Q.; Sriram, A.; Synnaeve, G.; and Collobert, R. 2020{\natexlab{b}}.
\newblock MLS: A Large-Scale Multilingual Dataset for Speech Research.
\newblock In \emph{Interspeech 2020}. ISCA.

\bibitem[{Radford et~al.(2023)Radford, Kim, Xu, Brockman, McLeavey, and Sutskever}]{radford2023robust}
Radford, A.; Kim, J.~W.; Xu, T.; Brockman, G.; McLeavey, C.; and Sutskever, I. 2023.
\newblock Robust speech recognition via large-scale weak supervision.
\newblock In \emph{International conference on machine learning}, 28492--28518. PMLR.

\bibitem[{Scannell(2007)}]{scannell12007crubadan}
Scannell, K. 2007.
\newblock The Cr{\'u}bad{\'a}n Project: Corpus building for under-resourced languages.
\newblock In \emph{Building and Exploring Web Corpora (WAC3-2007): Proceedings of the 3rd Web as Corpus Workshop, Incorporating Cleaneval}, volume~4, 5. Presses univ. de Louvain.

\bibitem[{Sutskever, Vinyals, and Le(2014)}]{NIPS2014_a14ac55a}
Sutskever, I.; Vinyals, O.; and Le, Q.~V. 2014.
\newblock Sequence to Sequence Learning with Neural Networks.
\newblock In \emph{Advances in Neural Information Processing Systems}, volume~27. Curran Associates, Inc.

\bibitem[{Taguchi and Chiang(2024)}]{taguchi-chiang-2024-language}
Taguchi, C.; and Chiang, D. 2024.
\newblock Language Complexity and Speech Recognition Accuracy: Orthographic Complexity Hurts, Phonological Complexity Doesn{'}t.
\newblock In \emph{Proceedings of the 62nd Annual Meeting of the Association for Computational Linguistics (Volume 1: Long Papers)}, 15493--15503. Association for Computational Linguistics.

\bibitem[{TAKAHASHI(1992)}]{takahashi1992kakasi}
TAKAHASHI, H. 1992.
\newblock KAKASI-Simple Kana Kanji Converter.
\newblock \emph{http://kakasi. namazu. org/}.

\bibitem[{Tang et~al.(2024)Tang, Yu, Sun, Chen, Tan, Li, Lu, MA, and Zhang}]{tang2024salmonn}
Tang, C.; Yu, W.; Sun, G.; Chen, X.; Tan, T.; Li, W.; Lu, L.; MA, Z.; and Zhang, C. 2024.
\newblock {SALMONN}: Towards Generic Hearing Abilities for Large Language Models.
\newblock In \emph{The Twelfth International Conference on Learning Representations}.

\bibitem[{Toshniwal et~al.(2018)Toshniwal, Sainath, Weiss, Li, Moreno, Weinstein, and Rao}]{toshniwal2018multilingual}
Toshniwal, S.; Sainath, T.~N.; Weiss, R.~J.; Li, B.; Moreno, P.; Weinstein, E.; and Rao, K. 2018.
\newblock Multilingual speech recognition with a single end-to-end model.
\newblock In \emph{2018 IEEE international conference on acoustics, speech and signal processing (ICASSP)}, 4904--4908. IEEE.

\bibitem[{Touvron et~al.(2023)Touvron, Lavril, Izacard, Martinet, Lachaux, Lacroix, Rozière, Goyal, Hambro, Azhar, Rodriguez, Joulin, Grave, and Lample}]{touvron2023llamaopenefficientfoundation}
Touvron, H.; Lavril, T.; Izacard, G.; Martinet, X.; Lachaux, M.-A.; Lacroix, T.; Rozière, B.; Goyal, N.; Hambro, E.; Azhar, F.; Rodriguez, A.; Joulin, A.; Grave, E.; and Lample, G. 2023.
\newblock LLaMA: Open and Efficient Foundation Language Models.
\newblock arXiv:2302.13971.

\bibitem[{Wang et~al.(2021)Wang, Riviere, Lee, Wu, Talnikar, Haziza, Williamson, Pino, and Dupoux}]{wang-etal-2021-voxpopuli}
Wang, C.; Riviere, M.; Lee, A.; Wu, A.; Talnikar, C.; Haziza, D.; Williamson, M.; Pino, J.; and Dupoux, E. 2021.
\newblock {V}ox{P}opuli: A Large-Scale Multilingual Speech Corpus for Representation Learning, Semi-Supervised Learning and Interpretation.
\newblock In \emph{Proceedings of the 59th Annual Meeting of the Association for Computational Linguistics and the 11th International Joint Conference on Natural Language Processing (Volume 1: Long Papers)}, 993--1003. Association for Computational Linguistics.

\bibitem[{Xie et~al.(2020)Xie, Luong, Hovy, and Le}]{9156610}
Xie, Q.; Luong, M.-T.; Hovy, E.; and Le, Q.~V. 2020.
\newblock Self-Training With Noisy Student Improves ImageNet Classification.
\newblock In \emph{2020 IEEE/CVF Conference on Computer Vision and Pattern Recognition (CVPR)}, 10684--10695.

\bibitem[{Xu, Baevski, and Auli(2021)}]{xu2021simpleeffectivezeroshotcrosslingual}
Xu, Q.; Baevski, A.; and Auli, M. 2021.
\newblock Simple and Effective Zero-shot Cross-lingual Phoneme Recognition.
\newblock arXiv:2109.11680.

\bibitem[{Yu et~al.(2024)Yu, Tang, Sun, Chen, Tan, Li, Lu, Ma, and Zhang}]{yu2024connecting}
Yu, W.; Tang, C.; Sun, G.; Chen, X.; Tan, T.; Li, W.; Lu, L.; Ma, Z.; and Zhang, C. 2024.
\newblock Connecting speech encoder and large language model for asr.
\newblock In \emph{ICASSP 2024-2024 IEEE International Conference on Acoustics, Speech and Signal Processing (ICASSP)}, 12637--12641. IEEE.

\bibitem[{Zhang et~al.(2023)Zhang, Han, Qin, Wang, Bapna, Chen, Chen, Li, Axelrod, Wang, Meng, Hu, Rosenberg, Prabhavalkar, Park, Haghani, Riesa, Perng, Soltau, Strohman, Ramabhadran, Sainath, Moreno, Chiu, Schalkwyk, Beaufays, and Wu}]{zhang2023googleusmscalingautomatic}
Zhang, Y.; Han, W.; Qin, J.; Wang, Y.; Bapna, A.; Chen, Z.; Chen, N.; Li, B.; Axelrod, V.; Wang, G.; Meng, Z.; Hu, K.; Rosenberg, A.; Prabhavalkar, R.; Park, D.~S.; Haghani, P.; Riesa, J.; Perng, G.; Soltau, H.; Strohman, T.; Ramabhadran, B.; Sainath, T.; Moreno, P.; Chiu, C.-C.; Schalkwyk, J.; Beaufays, F.; and Wu, Y. 2023.
\newblock Google USM: Scaling Automatic Speech Recognition Beyond 100 Languages.
\newblock arXiv:2303.01037.

\bibitem[{Zhao, Pratap, and Auli(2024)}]{zhao2024scalingsimpleapproachzeroshot}
Zhao, J.; Pratap, V.; and Auli, M. 2024.
\newblock Scaling A Simple Approach to Zero-Shot Speech Recognition.
\newblock arXiv:2407.17852.

\end{thebibliography}
\end{document}